# Citizen Participation and Machine Learning for a Better Democracy

MIGUEL ARANA-CATANIA

    University of Warwick, Coventry, and Alan Turing Institute, London, UK

FELIX-ANSELM VAN LIER

    Max Planck Institute, Halle, Germany

ROB PROCTER*

    University of Warwick, Coventry and Alan Turing Institute, London, UK

NATALIYA TKACHENKO

    University of Oxford, Oxford, UK

YULAN HE

    University of Warwick, Coventry, UK

ARKAITZ ZUBIAGA

    Queen Mary University of London, London, UK

MARIA LIAKATA

    Queen Mary University of London and Alan Turing Institute, London UK

*Corresponding author

**ABSTRACT**

The development of democratic systems is a crucial task as confirmed by its selection as one of the Millennium Sustainable Development Goals by the United Nations. In this article, we report on the progress of a project that aims to address barriers, one of which is information overload, to achieving effective direct citizen participation in democratic decision-making processes. The main objectives are to explore if the application of Natural Language Processing (NLP) and machine learning can improve citizens' experience of digital citizen participation platforms. Taking as a case study the "Decide Madrid" Consul platform, which enables citizens to post proposals for policies they would like to see adopted by the city council, we used NLP and machine learning to provide new ways to (a) suggest to citizens proposals they might wish to support; (b) group citizens by interests so that they can more easily interact with each other; (c) summarise comments posted in response to proposals; (d) assist citizens in aggregating and developing proposals. Evaluation of the results confirms that NLP and machine learning have a role to play in addressing some of the barriers users of platforms such as Consul currently experience.

**CCS CONCEPTS**

•Human-centered computing~Collaborative and social computing •Computing methodologies~Artificial intelligence~Natural language processing

**KEYWORDS**

Natural language processing, machine learning, digital citizen participation platforms

## 1. INTRODUCTION

The difficulties experienced by democratic institutions as agents of democratic will formation stem to some extent from an inability to truly embrace the potential of information and communication technologies (ICTs): almost every other aspect of life, including media, retail, tourism, activism etc., has been re-shaped, re-fashioned, re-formed by ICTs [41]. As a result, a gap between the activities of citizens and the way in which politics and democracy is carried out has developed, contributing to a decline in trust and confidence in governments [28], a decrease in participation [22] and a lack of innovative approaches to solve the complex and dynamic issues faced [74].

Direct democracy has been promoted as a way of countering this crisis in politics and studies have shown that it can lead to more informed people [9]; that it can be more effective in making decisions [86], e.g., by generating less debt [26]; that it better protects human rights [29], it improves basic services [16], etc. Now, with the growing availability of new digital technologies, governments are increasingly inviting a growing number of citizens to participate in policy-making [3, 62, 90]. Digital citizen participation platforms have been used to foster digital mass participation at different policy levels, ranging from participatory budgeting and local and national law-making to constitution-making [10, 42, 44, 48, 79].

Evidence suggests that the use of digital citizen participation platforms enhance the democratic legitimacy and inclusiveness, the quality of policy- and law-making, and the effectiveness of government [30, 61, 70]. If implemented effectively, such tools may be one way of addressing deteriorating trust and disillusion among the public with formal political institutions and parties [85]. In particular, online participation platforms may prove valuable for channelling increasingly individualised, ad hoc and single-issue-driven online political activism, which traditional, collective action-oriented organisations such as political parties and trade unions find difficult to absorb [88].

This article highlights information overload as a key practical challenge for crowdsourced policy- and law-making with digital tools and proposes ways to address it through the use of state-of-the-art NLP tools.

We begin with a review of the literature on digital platform-enabled citizen participation in policy making, documenting its growing popularity and how its objectives are compromised by a number of barriers, one of which is information overload. We then explain the background to the project and the methodology that we adopted in order to explore how the information overload problem might be tackled. We present the NLP and machine learning techniques that we have developed, followed by our methodology for evaluating their effectiveness. The results of the evaluation confirm that NLP and machine learning can help to tackle the information overload problem. We conclude with a summary of what we have learnt from the project and provide suggestions for further work if these techniques are to be widely adopted.

## 2. DIGITAL MASS PARTICIPATION IN LAW- AND POLICY-MAKING

Digital mass participation processes are generally understood as an open call for the public to participate via online tools in a law and policy-making process by sharing ideas, knowledge or opinions (see, e.g., [15]. Digital mass participation takes advantage of recent technological innovations to leverage large-scale collective thinking to make public decision-making more inclusive, more transparent, and improve its quality and legitimacy[1]. Participants are typically a self-selecting group of individuals who share information, knowledge, or talent and, in some cases, directly make decisions [4]. Digital mass participation processes run alongside existing representative institutions and inform and augment law- and policy-making processes with "more robust, frequent and disinterested advice-getting" [61, p. 363].

### 2.1. Digital Mass Participation, Collective Intelligence and Democracy

The underlying premise of digital mass participation is to improve public decision making by tapping into the collective intelligence of citizens [43, 61] or the "wisdom of the crowds" [45, 78]. Put simply, collective intelligence is based on the idea that "many heads are better than one". The theory posits that the aggregation of individual knowledge and opinions from large groups results in decisions that are often better than those made by any single member of the group or experts [78]. Although the concept itself is old and can be traced to antiquity [45], research on collective intelligence has more recently been fueled by the increasing pervasiveness of digital tools and the possibilities that an ever-growing number of people connected online offer.

Scholarship initially focused on how collaboration and open-source technology can be applied in the field of business and economics [36], for example, to help companies thrive [80] or enhance the effectiveness and efficiency of markets [77]; but it was quickly broadened to explore how it can help improve various other areas of society [57]. In particular, as technologies that facilitate more democratic participation have become more readily available, research on how collective intelligence can be employed to enhance democracy and improve the legitimacy of law and policy making has been rapidly expanding [5, 46, 50, 61, 70].

Aitamurto and Chen [2] usefully distinguish three distinct values that citizens' collective intelligence can bring to law- and policy-making. First, mass participation produces epistemic value. As mass participation widens the group of people involved in a law and policy making process beyond experts, it has the potential to generate knowledge that public officials would otherwise not have access to. Digital tools enable citizens to contribute experienced-based, "raw material" for policy makers [2], providing knowledge which is broader, more diverse and complete than expert knowledge and knowledge generated through public hearings or questionnaires [66]. That said, digital mass participation will also garner broader expert knowledge as such processes tend to engage professionals as well [15], but expert input will be blurred with the experienced-based knowledge of laymen. It is possible to target expert knowledge or particular communities by inviting only experts to participate [83]. However, the idea to use collective intelligence in broad-scale democratic processes is to counter traditional regulatory approaches that assume that only professional public servants possess the knowledge and skills to govern and to diversify knowledge and to pool together citizen's collective efforts and channelling it into law- and policy-making [61].

---

[1] Small scale deliberative forms of participation and mass participation are not mutually exclusive. For example, a small-scale randomly selected body can still solicit mass-input during its proceedings, as was done during the Irish Constitutional Convention.

Epistemic value is also created through peer-learning, as participants interact and exchange ideas on a participation platform. For example, a recent empirical study, which analysed the deliberative impact of mass participation in a Finnish crowdsourcing experiment showed that participants perceived the crowdsourcing exercise as educational [5]. Further, the study suggests that by facilitating "cross-cutting exposure" of participants with views that disagree with or had not previously considered, mass participation processes can become a deliberative space in which different arguments can be exchanged and discussed. The combination of these factors improves the quality of laws and policies and may lead to "more unbiased, independent and more accurate decisions than experts can." [66, p. 6].

Second, digital mass participation has democratic value, in that it improves the inclusiveness, transparency, and accountability of law- and policy making processes. Today, many policy and law-making decisions are nominally taken by representatives, but many tend to be meditated by often unelected, expert commissions. Digital mass participation opens up such processes to citizens. This does not necessarily mean that existing processes will be replaced, but they do open up new channels for participation that have the potential to significantly widen the range of citizens engaged in law and policy making. Further, digital mass participation also allows for more sustained citizen participation tools between election cycles [2], which means that citizens will be empowered, better informed, and more active in their political engagement [42, 53].

Digital mass participation improves the horizontal transparency of law and policy making, as citizens will be able to see the contributions and interact with other participants; as well as the vertical transparency between citizens and institutions, as citizens will have to be widely informed about the policy reform and be granted access to the civil servants and representatives responsible for the reform. This also improves the accountability of law and policy making, in that the public will be more informed about reform processes and have easier access to the responsible authorities [5]. In some areas, the use of participatory tools could bring about a "decentralisation of decision-making to the most effective and accountable unit" by empowering communities and enhancing their self-organization and social networking [66]. This type of power-shift has already been successfully practiced in public budgeting processes, where citizens are given control over a part of a community's budget to decide where best to invest public funds [75].

Finally, digital mass participation has an economic value. Given the access to a larger and more complete pool of knowledge and the inclusiveness of digital mass participation processes, policies and laws can be more effectively tailored around the people's needs, resulting in more effective policies that enjoy wider social support [50, 66]. It can also help lowering the costs of public good provision for governments, as the ideas generated through mass participation have the potential to garner innovative ideas at relatively low cost [2].

### 2.2. Digital Mass Participation in Practice

The idea of using collective intelligence via digital participation tools is rapidly gaining ground in the practice of law- and policy making as various factors challenge traditional models of top-down regulation.

First, societies in the digital era are increasingly complex. State institutions cannot rely anymore on traditional organisations and institutions of public life, such as political parties and unions, that previously functioned as "filters" and aggregators of public opinion [88]. Collective action is increasingly individualised, creating a 'new kind of politics' that is marked by a 'chaotic pluralism' that is too dynamic and too vast to be grasped or contained by traditional democratic processes [52]. As institutions are struggling to channel increasing individualised, ad hoc and single-issue-driven political discourse the gap between the citizens and the political class is widening. This is exemplified by the growing number of self-organised social mass movements that lack central organisational actors, peaking in global protest movements in 2019, in which a common theme was the lack of faith in traditional political systems and calls for more participatory democracy [89].

Second, internet penetration in developed countries has grown exponentially and information is readily available via the internet to everyone [49]. This has raised expectations among citizens about how governments are supposed to operate and how public services are to be provided, including increased demands for accountability, transparency and for more effective policies. To achieve effective law and policy making, government institutions will have to find new ways of gathering information to increase its capacity for evidence-based law making and for designing effective regulation that respond to the increasing complexity of governing [66, p. 12].

Third, the last 10 years have seen the development of increasingly sophisticated and user-friendly interactive technology to facilitate interactive participation of citizens, making such tools more accessible to a broader public. Originally predominantly used by grass-roots and protest movements, traditional state institutions are now exhibiting a greater willingness to embrace digital democracy tools. For example, in a recent report, the Council of Europe states that "opportunities arising from the new digital environment should be used to reinforce access to and participation in open culture, thereby strengthening democracy" [49]. Indeed, over the last five years or so state institutions have made increasing use of digital tools to foster citizen participation at different state levels [74], ranging from local and national law making [25, 38], to constitution making [63, 79].

### 2.3. Information Overload

The growth of citizen participation invariably leads to a new scale of digital public input data. High participation numbers are a crucial precondition for the effectiveness of digital mass participation: a too small set of participants constrains the collective intelligence function. At the same time, the increase of scale and complexity of citizen input can overwhelm both institutions and participants, putting at risk the value of digital mass participation [55, 72]. In a recent study, Chen and Aitamurto found that while policy makers show "strong intent to encourage civic engagement and utilize crowd input to help draft and revise policy", governments often lack the skills and resources to process citizen suggestions effectively [18]. Input from mass participation processes is often both too large in scope and too complex and content for policy makers to make sense of [18].

Often, the analysis of solicited citizen data only comes as an afterthought and is either processed manually or incompletely. For example, during the Chilean constitution making process annotators manually coded over 280,000 individual submissions in a very short time frame [27]. Similarly, the number of submissions in the Irish constitutional reform process was too large to be analysed and addressed systematically, so the constitutional convention decided to randomly draw individual submissions from the pool of submissions[2]. Likewise, users of a variety of e-democracy platforms have been found to experience difficulties with access and comprehension of online governmental information, following debates and processes, coordinating political action and collaboration, and duplicate contributions [24, 74, 82]. Research on online group communication has shown that "[t]he presence of too many participants [...] can turn online public spaces into noisy, overcrowded fora where no meaningful conversation can be held" [59].

In more technical terms, the difficulty of making sense of mass public input for decision makers and the "cacophony" that users experience can be described as information overload. The concept of information overload finds its roots in Herbert Simon's notion of 'bounded rationality', which recognises the limitations of "perfectly" rational decision making due to the cognitive limitations inherent in the human mind [73] (see [84]). Among these limitations is its reduced capacity to evaluate and process large amounts of information. Information overload, or simply "receiving too much information", produces cognitive biases as the human brain will routinely resort to heuristics in order to make decisions, which increase the likelihood of systematic errors in judgement [84]. The concept - sometimes discussed as cognitive overload [87], knowledge overload [39], or communication overload [7] - has been applied to a variety of contexts, but predominantly in the field of economics and management [23, 72]. More recently, the concept has been used to describe the challenges of online deliberation in democratic contexts [69] and digital citizen participation in law- and policy making [6, 18, 38, 51].

Information overload is a key challenge for digital mass participation, which jeopardises its intended democratic, epistemic, and economic values. For example, Information overload limits horizontal transparency as participants are not able "see and understand each other's" [1]. This lowers the capacity for meaningful exchange of ideas and learning and leads to a decline in the overall decline quality of mass participation. Information overload also limits digital mass participation's vertical transparency. Users of digital democracy platforms have found difficulties with access to and comprehension of online governmental information and in following debates and processes, thereby limiting the exchange of ideas, coordination of political action, and collaboration [24, 74, 82].

Political actors, in turn, will not expect to receive informed comments to consultative processes and may therefore tend to have a dismissive attitude to the contributions they receive [69]. Even if citizen input is of high quality, policy makers often lack the tools to meaningfully analyse and synthesise crowdsourced citizen input, which means that policy-makers are unable to make sense of and respond to citizens' contributions [18, 55, 72].

Crucially, the impact of digital participation platforms depends on political support [65]. When there is a lack thereof, participation remains 'trivial' and has no "little influence over outcomes" [30, 40]. Evidence shows that democratic innovations perceived as window dressing could lead to 'frustration, cynicism, or apathy' among citizens [30]. If tools are 'implemented poorly, citizen's expectations are not adequately managed, and decision makers do not act upon the results of digital consultations in a meaningful way', this participatory law and policy-making process may not only fail [3] but also impede the legitimacy of future participatory processes, and reinforce the crisis of legitimacy of democratic institutions [12, 69]. Taken together, information overload limits the effectiveness and the sustainability of long-term citizen participation and collaboration.

---
2 Personal communication: Interview with Irish Constitutional Convention Organiser. 2019.

Therefore, the value and success of digital mass participation depends on how effectively and meaningfully citizen input is solicited, analysed, and synthesized. Practitioners have started cursory explorations of using NLP tools to raise the capacity of such processes to organise a meaningful, communicatively 'complex deliberative process' and a discursive environment that is rich in terms of the categories of arguments and reasons it supports' [69] (see also [74]). Similarly, scholars are increasingly focusing on how to extract, classify, and analyze large-scale citizen-generated data, for example to better understand the nature of online public deliberation [54] or to deepen governments' knowledge of the citizen's views and needs [21]. Most studies in the field focus on the analysis of large-scale social media data [58, 76] and only a handful of analyses exist on data stemming from dedicated citizen participation processes, which tend to be limited in scope and scale [3, 33]. In addition, whilst valuable for refining and improving data analysis tools of large-scale citizen input for political decision makers, these studies tend not to address the issue of how to overcome the information overload experienced by citizens as a participatory process is ongoing. Improving the users experience of the process, however, is crucial for processes which seek to actively engage citizens. For digital citizen participation to be transparent and meaningful, they need to be accessible both to policy makers *and* citizens throughout the entire process. This paper remedies this gap by applying the use of NLP-tools in a large-scale scenario and by systematically testing how these tools can reduce the barriers that information overload poses both to citizens and decision makers.

## 3. THE PROJECT

The goal of the 'Citizen participation and machine learning for a better democracy project', funded by Nesta[3], is to evaluate the effectiveness of Natural Language Processing (NLP) methods as ways to tackle these barriers to the use of digital citizen participation platforms in democratic decision-making.

The platform chosen on which to deploy NLP methods is the open source platform Consul (www.consulproject.org). This is the largest digital democracy project worldwide, with more than 100 institutions involved, ranging from city administrations such as New York, Buenos Aires or Madrid, to national governments such as Colombia or Uruguay. It has been awarded the United Nations Public Service Award and is currently being used by the United Nations Development Programme and is also supported by the Inter-American Development Bank and the Open Government Partnership (see Figure 1).

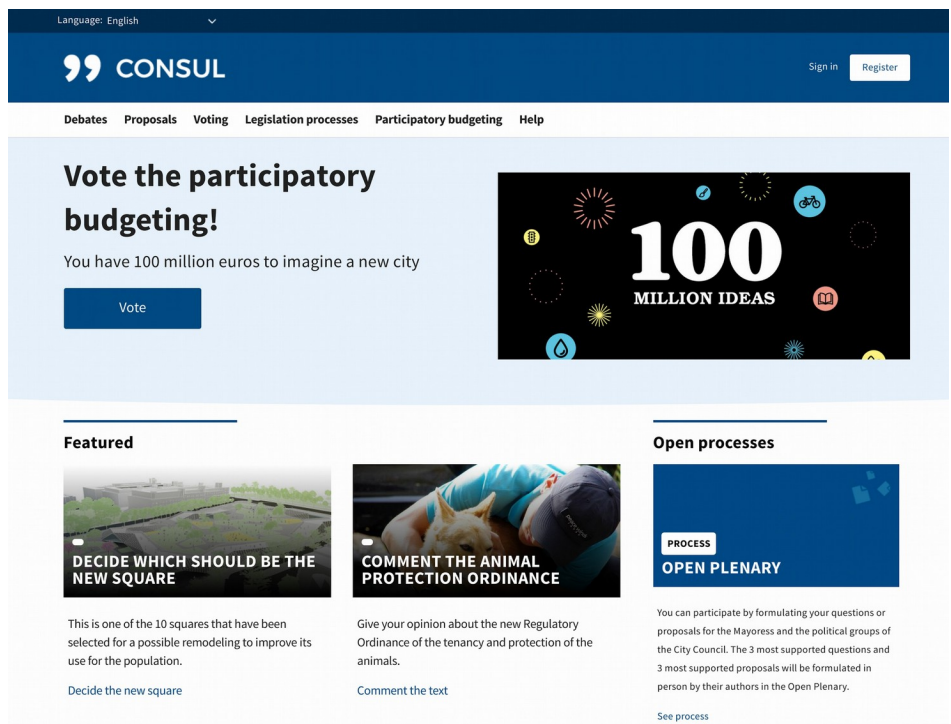

Figure 1: The Consul platform home page.

---

[3] https://www.nesta.org.uk/about-us/

The Consul platform is designed in a generic way to support collective intelligence processes, so that it can be used in many other settings not related to direct democracy where large-scale collective intelligence may bring benefits to institutions.

As with other digital mass participation processes, users of the Consul platform struggled with information overload. Both citizens and local government officials struggled to get an overview or summary of the many thousands of different proposals and comments created, so common objectives are difficult to achieve. For example, the use of Consul in Madrid, when considering specifically the 'Citizen proposals' process (see Figure 2) has resulted in more than 430,000 users submitting over 26,000 proposals and 125,000 comments and making more than 3,000,000 votes of support. However, in the last years only 2 proposals reached the minimum number of support votes (27,000) required for further action in this specific process. This skewed trend generalises to similar citizen participation platforms used by different governments, regardless of the local situation and their specific use cases.

Figure 2: An example proposal submitted to Consul, shown in the main English demonstration site of the project.

The project aims to test if the use of machine learning and NLP techniques on a digital platform for citizen participation will significantly increase the success of the processes that citizens can perform. As an example of the main participation process, work has been focused on citizen proposals, in which citizen users present proposals and must then obtain a certain number of support votes for their proposals to be selected for action by, e.g., local government administrators. With the use of carefully chosen machine learning and NLP techniques, we hypothesise that a greater number of proposals will reach the support needed to be selected.

State-of-the-art machine learning and NLP techniques have proven successful for facilitating access to large collections of information in related situations of collective sense-making. We argue that similar methods can be adapted to achieve the same effect and so overcome the critical barrier of information overload in citizen participation platforms. This will, in turn, allow these platforms to have a more profound impact on institutions and on the furthering of direct democracy.

Using these techniques, the Consul platform has been enhanced by the addition of information extraction and visualisation modules, to make it easier for users to use the platform and interact with each other and with the content within. The new modules have been designed to: a) categorize proposals to make it easier to aggregate and develop them; b) suggest relevant proposals for individual citizen users to support; c) group citizens so that they can interact with people with similar interests more easily; and d) summarise the comments to proposals to make it easier to understand people's opinions on their merits or disadvantages. Each of these modules are presented in detail in Sections 5 and 6.

The information extraction and visualisation modules have been implemented in such a way that they can be iterated and developed further by others. Consul is free software and any enhancements will be available to be reused and extended by all Consul installations.

## 4. DATASETS

The data used in the project has been the public participation datasets from the Consul instance of the Madrid City Council (the local instance is called 'Decide Madrid'). The datasets are available in open data format and have been downloaded from the 'Datos Abiertos'[4] repository, which provided files in the csv format, alongside metadata documentation, describing meaning and content of each file.

The case of Madrid has been chosen among the multiplicity of cases of the network of governments that use the Consul platform, not only because of the open publication of the data, but also because it represents one of the most successful cases of the whole network, having as we have mentioned almost half a million citizens registered. This has allowed us to work on a very large volume of proposals and comments to correctly evaluate the efficiency of the techniques developed to improve the problem of information overload. In addition, it is important to point out that the NLP and machine learning techniques we have developed depend only on the frequency of the words in the different documents and so are (in principle) independent of the language used. Hence, the main results are not limited to the Spanish context or the language chosen in our evaluation.

The dataset used contains 26,400 proposals, 125,135 comments, and 5,303 tags, all of them in Spanish. Both the analysis and processing of the data and the experiments have been done in Spanish, although in this article we will show some examples of proposals and tags translated into English for an easier understanding of them.

The main datasets of interest are Tags, Proposals and Comments. The Tags dataset contains a list of tags (i.e., a folksonomy) that have been manually entered by users of the Madrid platform and Proposals consists of the complete inventory of proposals submitted, including text (Title, Description and Summary fields), associated voting attributes (such as total votes, up-votes and down-votes) and Comments to the proposals.

Analysis of the tag usage revealed that the top 50 most used tags (from a total of 5,305; i.e. 0.9% of all tags) account for 80% of the most used tags across all proposals. It also confirmed one of the suspected deficiencies of the tagging functionality: the folksonomic freedom of using syntactic-morphologically distinct but semantically related language makes it more difficult for users to find proposals they might be interested in supporting.

Exploration of the Proposals dataset, which was used to generate new tags and to classify proposals into topical groups, was performed in order to establish some key parameters that will inform later work. For example, Figure 3 illustrates the distribution of word counts in the Title, Description and Summary fields of the Proposals dataset.

## 5. NLP METHODOLOGY AND TOOLS

In the following, we will describe the NLP techniques that we have used in this project. As stated earlier, our objective is to apply these techniques to reduce the information overload that is inevitable in a large dataset of posts, and to implement them directly on the participation platform so that they can be used in real time to improve the effectiveness of the participation processes. The techniques are focused on three main objectives: tag generation and proposal clustering, comments summarisation, and user clustering. These have been implemented to provide several new functionalities in the participation platform. In Sections 5 and 6 we describe this enhanced version of the platform, explain its use and the methodology we used to evaluate it.

---

[4] https://datos.madrid.es/sites/v/index.jsp?vgnextoid=644e11951b3bb510VgnVCM1000001d4a900aRCRD&vgnextchannel=374512b9ace9f310VgnVCM100000171f5a0aRCRD

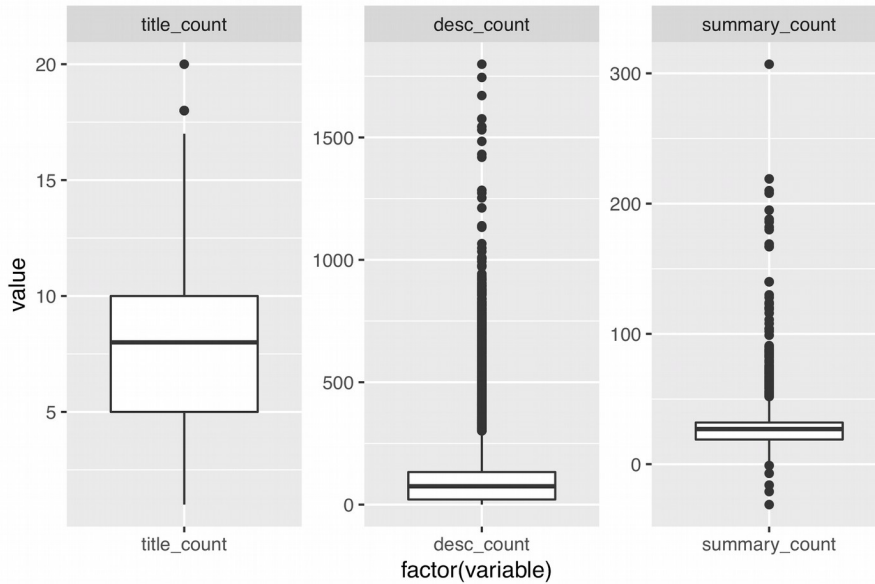

Figure 3. Statistical distribution of words in the Title, Description and Summary fields of the Proposals dataset.

### 5.1. Tag generation and proposal clustering

The categorisation of proposals has been carried out through topic modelling. The Title, Description and Summary fields were merged in a single, individual document for each proposal. These three elements are free text fields that are filled in very differently by each citizen, as can be seen in the very wide variability of the fields represented in Figure 3. In some cases, the Title contains the proposal itself, which is limited to one sentence, and the Description and Summary contain generic comments by the author on the proposal. In these cases, the most relevant terms characterising the proposal are therefore to be found in the Title of the proposal. In other cases, the Title has little content, and the authors elaborate in detail the proposal in the Description field. In these cases, the most relevant terms of the proposal are in this field. For this reason, in order not to lose the key terms necessary to categorise the proposal independently of the narrative style of each citizen, we decided to combine all the fields in a single text in order to carry out the analysis. Those texts were lemmatised using the StanfordNLP[5] library [71] (spaCy[6] and NLP-Cube[7] libraries were also tried but produced worse results for our Spanish dataset) with the AnCora corpora [81]. For the lemmatisation of the data, Part-of-Speech recognition was performed and noun forms selected. The texts were additionally cleaned removing stopwords through the NLTK[8] package [11], diacritic signs, URLs and HTML tags. Finally, the texts were tokenized, including the appropriate bi-grams.

The technique chosen on these texts for topic modelling was Non-Negative Matrix Factorization (NMF) [37, 47, 64]. The implementation was carried out using the scikit-learn[9] package [67], using the Frobenius Norm as the objective function, a Term-Frequency Inverse Document-Frequency (TF-IDF) representation of the words, Nonnegative Double Singular Value Decomposition [8, 14] for the initialization and a coordinate descent solver using Fast Hierarchical Alternating Least Squares [19, 31].

The main idea of NMF is to start from an initial representation of the analysed documents through a DxW size matrix that connects the D documents with the W words used in the whole collection of documents, and then factor this matrix into the product of two different matrices of sizes DxT and TxW. By making this split, a new intermediate T dimension is introduced that can be interpreted as the main topics of the documents. This factorisation also allows us to obtain a representation of the documents as a combination of topics (represented numerically by the DxT matrix), and a representation of each topic according to certain keywords (whose numerical representation is given by the TxW matrix).

NMF is a common topic modelling technique, but we also carried out an initial assessment of another very common technique is Latent Dirichlet Allocation (LDA) [13, 35]. NMF, as we have explained, iteratively searches for a matrix decomposition of the original matrix of documents into words. LDA models documents as generated by probabilistic distributions over topics and words, where, in particular, the probability of each topic in each document is given by a multinomial Dirichlet distribution [13, 35]. Both techniques are commonly used and may produce different results on a given dataset.

---

5 https://stanfordnlp.github.io/stanfordnlp/

6 https://spacy.io/

7 https://github.com/adobe/NLP-Cube

8 https://www.nltk.org/

9 https://scikit-learn.org

To compare the performance of the two techniques, we conducted a qualitative review of the keywords of the topics generated by each one on the dataset. First, we observed that LDA generated a significantly smaller number of distinct topics, such that when the number of topics was increased, the same topics and keywords were repeated in a redundant way. Second, following a careful examination of the dataset we were able to determine that NMF generated relevant topics that were not produced by LDA. For this reason, we chose the NMF technique. However, we do not present this as a general finding, to attempt which would be beyond the scope of this study. Rather, we conclude NMF to be better adapted to the characteristics of this particular dataset.

An analysis of the coherence and perplexity of the NMF generated topics was carried out to identify the optimal number of topics, but the results proved to be of little help. This was followed by a qualitative review of the results, where it was observed that approximately 40 topics was an optimal number, at which point the terms used to define the topics began to be repeated and therefore it was not useful to increase the number of topics further. This initial evaluation was carried out by comparing the proposals in the dataset with the topics obtained in order to identify whether the main topics referred in the proposal titles were also generated automatically through the topic modelling. In addition, this number was a good balance between, on the one hand, capturing the wide variety of topics in the platform and, on the other, being of most value when offering information to users. However, we should point out that the choice of this number of topics is only relevant to this particular dataset and therefore this value must be a free parameter to be modified in any implementation of this technique, so that the results can be evaluated in each case.

It is important to emphasize that when performing such an analysis, it is essential also to focus on the user experience and their interaction with the functionality. Our objective is to improve the effectiveness and quality of the participation processes. Therefore, the choice of techniques and the selection of parameters that affect them must take into account the implementation and final interaction of users with the platform. For example, in this case, this meant offering users a large enough number of topics to allow sufficient deliberative richness, while not limiting them to a small number of topics, regardless of whether this might seem equally valid from a technical point of view when assessing the factoring of documents into topics and terms. In Section 6 we explain the implementation of the topic modelling and the other techniques to illustrate how they provide new functionalities that are simple and intuitive for users.

As mentioned above, one of the outcomes of the NMF analysis is the representations of proposals in the topic space as a linear combination of the different topics. This enables identification of the proposals that best represent each of the topics.

Below, we present as an example our English translation of the titles of the three most representative proposals for some of the topics:

*Topic 5 [terms: underground line bus lift stop]*

    New underground line in the east of Madrid

    Public bicycles in Almudena underground

    Underground in Valderrivas neighbourhood

*Topic 7 [terms: resident car parking car_park]*

    Outlets in the park and ride facilities

    Remove unused cars from parking lots

    Exclusive parking for electric cars on each street

*Topic 9 [terms: pet animal owner cat excrement]*

    For the Rights of the animals

    Improved protocol for identification of lost animals

    Public hospital for animals and pets

*Topic 30 [terms: housing rent price flat tax]*

    Improvement in renting and buying housing

    Social housing

Demands of Housing Groups in the City of Madrid

The topic modelling allowed us to generate 40 topics, each defined by 5 terms, i.e., a total of 200 terms. For each proposal this also tells us how representative each of these 200 terms is of the content of the proposal. We decided to select for each proposal the 5 most representative terms and make these the tags of the proposal. This way we automatically obtained tags for each proposal.

The usual approach to identifying similar proposals is to cluster proposals according to their similarity. In this case, however, our objective is to provide a list of most similar proposals for each proposal. Therefore, we are not interested in clusters of proposals but in the distance for each proposal with respect to all the others, based on the metric used for the clustering. Since we have obtained the coefficients that define each proposal in the topic space, each proposal has a position defined by those coefficients, and thus we can calculate the distance between any two proposals as the Euclidean norm of the vector that connects their positions. For each proposal we collected the 50 closest proposals sorted by distance.

Below we show as an example our English translation of two proposals, together with the 5 proposals closest to them:

*Proposal Title: Banning advertisements on cars*

- Prohibition of advertising flyers on car windshields

- Windscreen advertisement

- An end to SPAM on paper

- Eradicate windshield wiper advertising

- Prohibition of advertising on any type of car windshield

*Proposal Title: Closing downtown traffic to non-resident vehicles*

- Prohibition of single-occupant vehicles

- Priority Residential Area in Center

- Collection of fees from vehicles circulating in and around Madrid

- Restricting private vehicle traffic on Gran Via road on Sundays and public holidays

- Extend regulated parking to the entire municipality

### 5.2. Comments summarisation

The comments associated with each proposal were combined, producing a single text for each proposal to then be summarised.

We considered initially two models of extractive summarisation. The first model consisted of giving each sentence a score by adding the values of the terms that make it up according to a TF-IDF model, selecting the sentences with the highest score as a summary. The TF-IDF model represents each term numerically by using two values: the frequency with which the term appears in the document under consideration (this part is referred as TF for Term Frequency), and the inverse frequency with which the term appears in all the documents (hence the designation IDF for Inverse Document Frequency). By multiplying these two values, the relevance of each term is obtained, aiming for the term to be very frequent in the selected document but not frequent in other documents, thus especially relevant to the subject under discussion. This is a simple way of measuring the relevance of terms used, which can be used independently of the language in question and therefore would be an advantage in a globally used platform.

The second model, TextRank, begins with the representation of sentences as vectors from the SBWC's GloVe embeddings [17, 68] (Spanish Billion Word Corpus[10]). In the second step, a network is created for each proposal where each sentence represents a node and the distance between nodes is given by the cosine similarity between the vectors. In the third final step of this model, the PageRank algorithm, implemented through the NetworkX[11] package [32], was applied to each network to obtain the main nodes and, hence, the most relevant sentences of each proposal [34, 56]. The TextRank model is not immediately reusable in other languages, due to the requirement for embeddings trained in a particular language, however, this allows capturing a greater richness of the language used. Therefore, we chose to use this model in the subsequent steps of the project.

---

10 https://crscardellino.github.io/SBWCE/

11 https://networkx.github.io/

### 5.3. User clustering

As stated above the aim of this module is to connect users who may have common interests to facilitate their collaboration.

The number of proposals and comments created by each active user (users that have created either proposals or comments in the platform) were analysed. This information is represented in Figure 4. It can be seen that the majority of users (~65%) created a single proposal or a comment, and the majority of the remainder created up to 5 proposals and up to 10 comments.

The first method tried was to cluster users taking into account the similarities in the list of proposals on which they commented. However, as seen in the previous figure, this is not enough to establish a relationship between users. For example, it was observed that 84% of the users only have a single interaction with other users.

The chosen method for user clustering was NMF topic modelling on texts produced by merging all the content created by each user. That enabled us to obtain the distance between texts in the topics space and thus the similarity between users defined by this distance. In this regard, it should be noted that the connection between users is only made between those who have created content, and only depends on content that is public. This is an advantage in terms of privacy compared to what is usual in commercial social networks, since it does not use users' private information about what proposals they have supported.

In relation to the same method used in section 5.1 for the topic modelling of proposals, it is important to note a difference between the two. In that case each proposal is an independent element, since the objective is to organise the proposals themselves, while in user clustering all the content produced by a user is merged, since the objective is to organise the users themselves.

It should be noted that in NMF topic modelling the identification of similar elements depends only on the shared use of terms relevant to each topic. Therefore, users are grouped together as their comments focus on common topics, but whether users have different or similar views on the topic concerned is not distinguished, only that they are interested in the same topics. This is an advantage, as it encourages users with different positions to debate on topics of common interest, rather than favouring the creation of opinion bubbles as in some popular social networks.

As we have mentioned, most users post few proposals or comments. If we were to consider only the case of users who have posted a single proposal, clustering would essentially be reduced to the previous clustering of proposals. That is, we would connect users with other users who have posted proposals on the same topic. However, for users who have posted more than one proposal the connection here will be multiple, mixing the different topics they have dealt with in their proposals. And more interestingly, the current scenario also considers other cases, such as users who have only commented on proposals. It is able, for example, to connect a user who has posted a single proposal about parks with another user who has mentioned the park concept in a comment to any other proposal in the platform. In this sense, the topic modelling technique does not particularly favour the length of the texts, but rather the relevance of the terms used, as we explained when describing the TF-IDF technique.

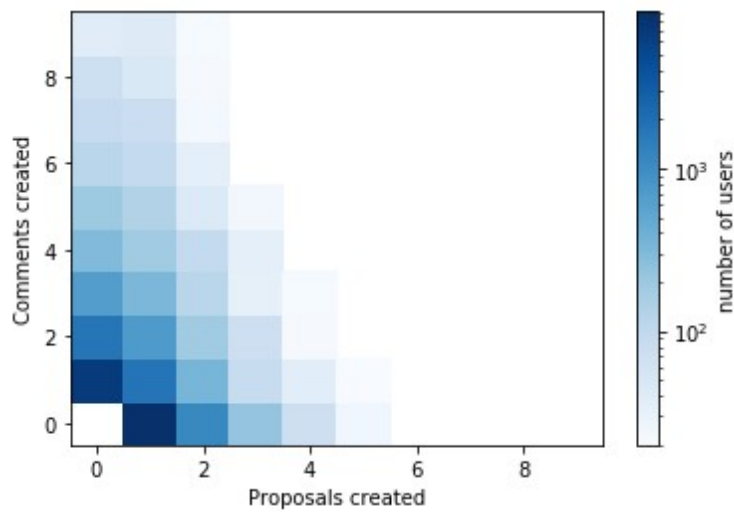

Figure 4: Number of comments and proposals created by active users.

## 6. EXPERIMENTAL EVALUATION

### 6.1. Design

The evaluation of the new NLP-based information retrieval tools was carried out using a lab-based design. It was held at Medialab-Prado[12], a digital cultural centre in Madrid. 14 subjects were recruited through an open call on Medialab-Prado's website and social networks. The sample is therefore not intended to be statistically representative of the entire population of the city, but to serve as a first assessment of the impact of the techniques developed. The results should be considered within that scope and taking into account the statistical margins that we will offer throughout this section.

None of the subjects were involved in the development of these techniques, nor did they know anything about them prior to the evaluation. In addition, as will be described in more detail below, the fact that the evaluation compared different techniques was unknown to the subjects, as the system switched between these techniques without providing any identifying details.

The evaluation design consisted of users performing a series of tasks using the Consul platform, where the effectiveness of the information retrieval tools and the original platform were compared. Two versions of the Consul platform were used for the evaluation: the original version (condition (a)) and an enhanced version in which the NLP tools have been integrated (condition (b)). The enhanced Consul platform is shown in Figure 5.

The evaluation followed a *within subjects* design [60], where subjects were allocated to conditions (a) and (b) and were given a series of tasks to complete. The two conditions were randomised for each subject to counter learning effects. The aim was to test the hypothesis that information retrieval is more effective in condition (b). Effectiveness was measured by selected dependent variables, e.g., (i) how long subjects took to complete a task; (ii) subjects' assessment of the quality of the task results they obtained. Differences in the dependent variables for each condition were tested for statistical significance. Subjects were also asked to complete a short questionnaire to assess their opinions of the version of Consul they used each time. Questionnaires were used to measure subjective satisfaction.

The tasks, which are described in more detail below, were designed to test the following use cases, which following a review, we determined to be prototypical given the objectives of citizen participation platforms such as Decide Madrid:

1. A user is looking for one or more proposals they would like to support. To do this, they choose one or more tags in successive steps until they find four proposals that are similar.
2. A user wants to find proposals that are similar to the one they have already found. To do this, they use the related proposals lists until they find four proposals that are similar.
3. A user wishes to decide which proposal they would like to support by reading a summary of comments others have left.
4. A user wants to find like-minded citizens. To do this, they use the related users lists until they find four users that are similar.

Post-test, subjects were interviewed to: (a) assess their opinions of the task; (b) provide an opportunity to triangulation the quantitative results and; (c) acquire more general comments about the evaluation.

Tasks (see below) were designed to measure a selection of dependent variables chosen to reflect the goal of each use case.

*6.1.1 Use Case One: Finding the Most Similar Proposals (1)*

Subjects in both conditions were given a specific proposal and instructed to use the tags facility to find 4 similar proposals. Dependent variables: (i) time taken; (ii) subjects' assessment of the match of the proposals retrieved; and (iii) subjects' assessment of the ease of the task. This was repeated 6 times (3 with the 'original' platform and 3 with the 'enhanced' platform). The proposals were selected from 6 different topics. The proposals were the most representative proposals of the topic.

The subjects were split in two groups and the same proposals were swapped between the two conditions to reduce as much as possible the factors affecting the evaluation. The order in which the subjects used the two versions was randomised in a within subjects design to eliminate training effects.

*6.1.2 Use Case Two: Finding the Most Similar Proposals (2)*

Subjects in the enhanced Consul condition were given a specific proposal and asked to use the related proposals facility on the Consul platform to find 4 similar proposals. Subjects were asked to rate the relevance of the proposals retrieved. Dependent variable: (i) time taken; (ii) subjects' assessment of the relevance of proposals retrieved to the original proposal; and (iii) subjects' assessment of the ease of the task. This was repeated 5 times. The proposals were selected from 5 different topics. The proposals were the most representative proposals of the topic.

---

12 https://www.medialab-prado.es/en

*6.1.3 Use Case Three: Summarising Comments*

Subjects in the enhanced Consul condition were given a specific proposal. They then were asked to read the comments posted in response to the proposal and then evaluate the quality of the summary provided. Dependent variable: (i) subject's satisfaction with the summary; and (iii) subjects' assessment of the ease of the task. This was repeated 5 times. The proposals were selected: as 2 proposals with a small number of comments (10 comments); 2 proposals with a medium number of comments (50 comments); and 1 proposal with a large number of comments (between 450 and 1000 comments).

Figure 5: The Consul platform with enhanced tagging and clustering of proposals.

*6.1.4 Use Case Four: Finding Like-Minded Citizens*

Subjects in the enhanced Consul condition were given a specific user. They then were instructed to find 4 citizens with similar interests from the related users list. Dependent variables: (i) time taken; (ii) subjects' assessment of the quality of the results; and (iii) subjects' assessment of the ease of the task. This was repeated 5 times. The users were selected from 5 different groups and were identified as being the most representative of the groups.

## 7. RESULTS

We reproduce below the results for each task.

### 7.1 Task 1: Search for similar proposals using tags

We observed a reduction in mean time of 40.9% when using the enhanced version (see Table 1). The reduced number of subjects implies a very limited statistic to corroborate the results, but points to a clear pattern of improvement that we can see reproduced also for each subject individually. A Student's t-test confirmed that the results were statistically significant at $p<0.0003$.

For similarity of proposals, the mean results were 3.65/3.95 (original/enhanced), an improvement although not statistically significant. For ease of performing the task, the results were 3.88/4.32 (original/enhanced), significant ($p<0.02$).

Examining the results of each proposal individually, we found the original version sometimes performed better than the enhanced version. These were proposals in which the author used very defined and descriptive labels, allowing the search to be made on a much smaller set of proposals. As we noted before this is very uncommon: the 50 most used tags, which represent 80% of all tags used, are essentially generic terms such as "environment" or "urbanism". We also note that when using the original version of the platform, subjects could sometimes find only 2-3 similar proposals.

Table1. Search for similar proposals using tags.

|  | Original Mean | Enhanced Mean | Original SD | Enhanced SD | p |
|---|---|---|---|---|---|
| **Time (s)** | 381 | 225 | 216 | 178 | 0.0003 |
| **Similarity (1-5)** | 3.65 | 3.95 | 1.0 | 1.09 | 0.1 |
| **Ease (1-5)** | 3.88 | 4.32 | 1.04 | 0.91 | 0.02 |

In post-test interviews, subjects mentioned that in some cases, referring to the original version, labels were too few or too broadly defined, and so were not very useful for finding similar proposals.

### 7.2. Task 2: Search for similar proposals using lists of related proposals

We observe again a reduction in mean time, in this case of 58,7%, and this is statistically significant ($p<0.00000005$) (Table 2). We also observe a higher similarity ($p<0.06$) and higher ease of use ($p<0.02$) rating. In post-test interviews, all subjects described this task carried out with the enhanced version as much simpler and the proposals more clearly related to each other than in the first task.

Table 2. Search for similar proposals using lists of related proposals.

|  | Original Mean | Enhanced Mean | Original SD | Enhanced SD | p |
|---|---|---|---|---|---|
| **Time (s)** | 381 | 157 | 216 | 74 | 0.00000005 |
| **Similarity (1-5)** | 3.65 | 3.95 | 1.0 | 0.86 | 0.06 |
| **Ease (1-5)** | 3.88 | 4.27 | 1.04 | 0.85 | 0.02 |

### 7.3. Task 3: Summarisation of comments

The average relevance result of 3.11 suggests that the relevance of the extracts was not clear (Table 3). In post-test interviews, subjects mostly mentioned the disparity between the extracts, some very well selected and others very poorly selected. They also mentioned that, in some cases, a selection of sentences did not add meaningful information, making clear the need for improvement.

Table 3. Summarisation of comments.

|  | Enhanced Mean | Enhanced SD |
|---|---|---|
| **Relevance (1-5)** | 3.11 | 1.21 |
| **Ease (1-5)** | 4.17 | 0.80 |

### 7.4. Task 4: Search for similar users

The objective of this task is very similar to that of task 2. Comparing times, we observe an increase in time of 29%, so although this task was more complicated to perform, the results are sufficiently alike to suggest that finding similar users is like finding similar proposals, indicating the benefit of this new functionality (Table 4). Some difference is to be expected since the concept "similar proposal", although subjective, as the subjects themselves remarked, is more defined than the concept "similar user", since for each user multiple proposals and comments must be taken into account for the comparison.

Again, the results indicate a positive trend on the relevance of the task, although they seem to confirm that this task is more complicated than searching for proposals.

Table 4. Search for similar users.

|  | Enhanced Mean | Enhanced SD |
|---|---|---|
| **Time (s)** | 202 | 135 |
| **Similarity (1-5)** | 3.76 | 1.10 |
| **Ease (1-5)** | 3.63 | 1.30 |

In the post-test interviews, several subjects mentioned that the criterion of similarity between users was not as clear as similarity between proposals. They also confirmed the difficulty of using this functionality, which was partly due to anonymization of usernames, which homogenised the lists

### 8. DISCUSSION

From the results of the evaluation, we can conclude that the NLP and machine learning techniques we developed improved the effectiveness of citizen participation and collective intelligence processes in a significant way.

First, the time required to carry out the search tasks for similar proposals with the enhanced version of Consul was found to be approximately 40% to 60% less. Subjects also assessed these tasks to be considerably easier to carry out with the enhanced version. These quantitative findings were corroborated by comments made by subjects when they were interviewed post-test. In some cases, not only was there a relative improvement in task performance, but as some users pointed out, it was impossible to find similar proposals using the original version of Consul. Therefore, we not only found a performance improvement with the enhanced version but that it also provided new functionality.

Second, subjects repeatedly pointed out that the categorisation of proposals with the enhanced version of Consul was better and, in addition, it allowed the inclusion of a greater number of relevant categories for each proposal. This is important not only in terms of improving the performance of this particular task, but in a more general way it means that the proposals were now structured in a more accessible way in the enhanced version. Whereas, for administrators, the original version of the platform only allows the detection of relevant issues through the number of supports of some successful proposals, the enhanced version would enable a global analysis of posts and extraction of information such as the most relevant topics for citizens and the different proportion of contributions for each topic, etc.

Third, the evaluation results show how the introduction of the NLP and machine learning techniques enabled tasks that previously were not feasible, such as the summarisation of texts or the discovery of users with similar interests. Both were also valued positively by subjects.

Fourth, the results showed that the time required in the task of identifying similar users was similar to that observed for the task of identifying similar proposals when using the enhanced version. This reinforces the conclusion that this new mode of interaction would enhance users' experience of the platform. The NLP and machine learning techniques enabled types of interactions between users and their ideas that were not possible before. This suggests that an improvement in relation to the collective intelligence of the user community is in prospect, which would enable new ways of thinking collectively. We foresee opportunities for further development in this regard.

Finally, although the four tasks selected for the evaluation are quite specific, they also comprise the most essential types of interaction on citizen participation platforms of this kind. In addition, these types of interactions are also key to more complex participation processes, such as collaborative legislation processes or participative budgeting. This implies that the improvements we have observed would have indirect consequences for the effectiveness of participation processes far beyond those explored in this project.

## 9. CONCLUSIONS AND FUTURE WORK

In this article we have reported findings from a project to develop NLP and machine learning tools to tackle the problem of information overload in digital citizen participation platforms. Though small in scale, the evaluation results confirm at a significant level our hypothesis that these tools can make certain key tasks easier for users of these platforms.

Current digital participation platforms rely entirely on their users for the generation and processing of content, which, as we have argued, significantly limits their effectiveness. This project has demonstrated the possibility of surpassing the existing effectiveness ceiling, thereby greatly increasing the social impact of digital participation platforms and opening up the possibility of attacking much more complex collective intelligence problems in society. Hence, we believe that we have succeeded in laying a foundation for future work to improve and build on these initial achievements.

With this in mind, one of our next objectives will be to improve the performance of the text summarization function. As can be seen from the evaluation results, there is room for improvement compared to the other tasks. At the same time, there are a range of techniques for automatic text generation suitable for multi-language use (e.g., [20]) that with a different approach, may allow for a considerable improvement in performance and we plan to investigate these.

We plan to carry out further evaluations to expand the evidence base for the value of NLP and machine learning tools in digital participation platforms. These will include experimental evaluations using the same methodology as we have already carried out, but with larger numbers of subjects and additional new techniques in the different tasks, such as it would be in the case of the summarization task we have mentioned above.

However, experimental evaluations can only go some way to establishing whether the new tools we have developed will succeed in creating a better experience for users of digital participation platforms like Consul. Our aim now will be to conduct a live evaluation of the Consul platform augmented with the NLP-based and machine learning techniques. One possible approach would be to perform out an A/B test on a public version of the platform, where users of the platform interact with one of two versions (A or B) of the platform, one of which includes the new NLP and machine learning modules and the other is the standard version. This evaluation could be carried out through a random selection of incoming users, possibly stratified by demographic features to ensure a sample as representative as possible of the population. Although, in the current case, the evaluation has been carried out on citizen proposals, this new study could be extended to other citizen participation processes, such as participatory budgeting, which share a similar participatory process design and the same problems of information overload, so we hope to demonstrate that the techniques we have developed will be just as effective. We are currently in discussions with local government authorities in the UK about their participation in a live evaluation.

Beyond improving the functionality of NLP and machine learning tools, more work needs to be done to ensure the trust and their social acceptance. Given the sensitivity of democratic processes, digital democracy tools need to be designed to be in compliance with highest ethical and legal standards and with an eye on new forms of potentially harmful political discourse and manipulation. As of now, too little thought has been given to questions of systematic and comprehensive process design, which would need to include the elaboration of auditing and impact assessment frameworks to monitor the construction and scrutinize the internal workings of algorithms and their conformity with existing rules and regulations (e.g., privacy protection and equalities legislation) and their potential to produce biased results prior to their implementation; and monitor their broader societal impact once employed.

**ACKNOWLEDGMENTS**

This project was funded by Nesta under its Collective Intelligence programme[13] and by the Alan Turing Institute for Data Science and Artificial Intelligence. We would like to thank our Nesta colleagues for their support. We would like to thank Medialab-Prado for their support in the realization of the experiments. We would also like to express our appreciation to the citizens of Madrid who generously gave of their time to contribute to the project.

---

13 https://www.nesta.org.uk/feature/collective-intelligence-grants/

**REFERENCES**

1. Tanja Aitamurto. 2016. Collective intelligence in law reforms: When the logic of the crowds and the logic of policymaking collide. In 2016 49th Hawaii International Conference on System Sciences (HICSS), pp. 2780-2789.
2. Tanja Aitamurto, and Kaiping Chen. 2017. The value of crowdsourcing in public policymaking: epistemic, democratic and economic value. *The theory and practice of legislation* 5.1: 55-72.
3. Tanja Aitamurto, Kaiping Chen, Aitamurto, Tanja, Kaiping Chen, Ahmed Cherif, Jorge Saldivar Galli, and Luis Santana. 2016. Civic CrowdAnalytics: Making sense of crowdsourced civic input with big data tools. In Proceedings of the 20th International Academic Mindtrek Conference, pp. 86-94.
4. Tanja Aitamurto, T., Hélène Landemore, and Jorge Saldivar Galli. 2017. Unmasking the crowd: participants' motivation factors, expectations, and profile in a crowdsourced law reform. *Information, Communication & Society*, *20*(8), 1239-1260.
5. Tanja Aitamurto, and Hélène Landemore. 2015. Five design principles for crowdsourced policymaking: Assessing the case of crowdsourced off-road traffic law in Finland. Journal of Social Media for Organizations, 2, 1 (2015), 1-19.
6. Zach Bastick. 2017. Digital Limits of Government: The Failure of E-Democracy. In *Beyond Bureaucracy*, pp. 3-14. Springer, Cham.
7. Joao Carlos Lopes Batista, and Rui Pedro Marques. 2017. An overview on information and communication overload. In Information and communication overload in the digital age (pp. 1–19). IGI Global.
8. Mark Belford, Brian Mac Namee, and Derek Greene. 2018. Stability of topic modeling via matrix factorization. *Expert Systems with Applications*, *91*, 159-169.
9. Matthias Benz, and Alois Stutzer. 2004. Are voters better informed when they have a larger say in politics?. Evidence for the European Union and Switzerland. *Public choice* 119, no. 1-2: 31-59.
10. Carlos Bernal. 2019. How Constitutional Crowd-drafting can enhance Legitimacy in Constitution-Making. In H. Lerner & D. Landau (Eds.), Comparative Constitution Making. Edward Elgar.
11. Steven Bird, Ewan Klein, and Edward Loper. 2009. Natural language processing with Python: analyzing text with the natural language toolkit. O'Reilly Media, Inc.
12. Ricardo Blaug. 2002. Engineering democracy. Political studies, 50, 102-116.
13. David Blei, Andrew Ng, and Michael Jordan. 2003. Latent dirichlet allocation. Journal of machine Learning research, 3(Jan), 993-1022.
14. Christos Boutsidis, and Efstratios Gallopoulos. 2008. SVD based initialization: A head start for nonnegative matrix factorization. *Pattern recognition*, *41*(4), 1350-1362.
15. Daren Brabham. 2013. Crowdsourcing. The MIT Press.
16. Yves Cabannes. (2015). The impact of participatory budgeting on basic services: municipal practices and evidence from the field. Environment and Urbanization. 27. 257-284. 10.1177/0956247815572297.
17. Cristian Cardellino. 2016. Spanish billion words corpus and embeddings. *Spanish Billion Words Corpus and Embeddings*.
18. Kaiping Chen, and Tanja Aitamurto. 2019. Barriers for Crowd's Impact in Crowdsourced Policymaking: Civic Data Overload and Filter Hierarchy. *International Public Management Journal* 22, no. 1: 99-126.
19. Andrzej Cichocki, and Anh-Huy Phan. 2009. Fast local algorithms for large scale nonnegative matrix and tensor factorizations. *IEICE transactions on fundamentals of electronics, communications and computer sciences*, *92*(3), 708-721.
20. Jacob Devlin, Ming-Wei Chang, Kenton Lee, and Kristina Toutanova. 2018. Bert: Pre-training of deep bidirectional transformers for language understanding. *arXiv preprint arXiv:1810.04805*.
21. Olfa Driss, Sehl Mellouli, and Zeineb Trabelsi. 2019. From citizens to government policy-makers: Social media data analysis. Government Information Quarterly, 36(3), 560–570.
22. John Dryzek, Andre Bachtiger, Simone Chambers, Joshua Cohen, James Druckman, Andrea Felicetti, James Fishkin, et al. 2019. The crisis of democracy and the science of deliberation. *Science* 363 (6432): 1144-1146.
23. Martin Eppler, and Jeanne Mengis. 2004. The Concept of Information Overload—A Review of Literature from Organization Science, Accounting, Marketing, MIS, and Related Disciplines. The Information Society, 20(5), 1–20
24. Enzo Falco, and Reinout Kleinhans. 2018. Beyond technology: Identifying local government challenges for using digital platforms for citizen engagement. *International Journal of Information Management* 40: 17-20.
25. Cristiano Faria, and Malena Rehbein. 2016. Open parliament policy applied to the Brazilian Chamber of Deputies. The Journal of Legislative Studies, 22(4), 559–578.
26. Lars Feld, and John Matsusaka. 2003. Budget referendums and government spending: evidence from Swiss cantons. *Journal of Public Economics* 87, no. 12: 2703-2724.
27. Constanza Fierro, Claudio Fuentes, Jorge Pérez, and Mauricio Quezada. 2017. 200K+ Crowdsourced Political Arguments for a New Chilean Constitution. Proceedings of the 4th Workshop on Argument Mining, 1–10.
28. R. S. Foa, Klassen, A., Slade, M., Rand, A., Collins, R. 2020. The Global Satisfaction with Democracy Report 2020. Bennett Institute for Public Policy, University of Cambridge.
29. Bruno Frey, and Lorenz Goette. 1998. Does the Popular Vote Destroy Civil Rights? American Journal of Political Science, 42(4), 1343.
30. Archon Fung. 2015. Putting the Public Back into Governance: The Challenges of Citizen Participation and Its Future. Public Administration Review, 75(4), 513–522.
31. Nicolas Gillis. 2014. The why and how of nonnegative matrix factorization. *Regularization, optimization, kernels, and support vector machines*, *12*(257), 257-291.
32. Aric Hagberg, Pieter Swart, and Daniel Chult. 2008. *Exploring network structure, dynamics, and function using NetworkX* (No. LA-UR-08-05495; LA-UR-08-5495). Los Alamos National Lab. (LANL), Los Alamos, NM (United States).
33. Loni Hagen, Teresa Harrison, Ozlem Uzuner, Tim Fake, Dan Lamanna, and Christopher Kotfila. 2015. Introducing textual analysis tools for policy informatics: A case study of e-petitions. Proceedings of the 16th Annual International Conference on Digital Government Research, 10–19.

34. Kazi Saidul Hasan, and Vincent Ng. 2010. Conundrums in Unsupervised Key phrase Extraction: Making Sense of the State-of-the-art. In Proceedings of the 23rd International Conference on Computational Linguistics: Posters (pp. 365-373). https://www.aclweb.org/anthology/C10-2042.pdf
35. Matthew Hoffman, Francis Bach, and David Blei. 2010. Online learning for latent dirichlet allocation. In advances in neural information processing systems (pp. 856-864).
36. Jeff Howe. 2006. The rise of crowdsourcing. Wired magazine, 14, 6, 1-4.
37. Patrick Hoyer. 2004. Non-negative matrix factorization with sparseness constraints. *Journal of machine learning research*, *5* (Nov), 1457-1469.
38. Yu-Tang Hsiao, Shu-Yang Lin, Audrey Tang, Darshana Narayanan, and Claudina Sarahe. 2018. vTaiwan: An Empirical Study of Open Consultation Process in Taiwan [Preprint]. SocArXiv
39. Richard Hunt, and Richard Newman. 1997. Medical knowledge overload: A disturbing trend for physicians. Health Care Management Review, 22(1), 70–75.
40. Renee Irvin, and John Stansbury. 2004. Citizen Participation in Decision Making: Is It Worth the Effort? Public Administration Review, 64(1), 55–65.
41. Rob Kitchin. 2014. The real-time city? Big data and smart urbanism. *GeoJournal.*79: 1-14.
42. Karolina Koc-Michalska, and Darren Lilleker. 2017. Digital politics: Mobilization, engagement, and participation: 1-5.
43. Hélène Landemore. 2017. Democratic reason: Politics, collective intelligence, and the rule of the many. Princeton University Press.
44. Hélène Landemore. 2015. Inclusive constitution-making: The Icelandic experiment. *Journal of Political Philosophy* 23, no. 2: 166-191.
45. Hélène Landemore, and Jon Elster (Eds.). 2012. Collective Wisdom. Cambridge University Press.
46. Maria Lastovka. 2015. Crowdsourcing as new instrument in policy-making: making the democratic process more engaging. European View, 14, 1, 93-99.
47. Daniel Lee, and Sebastian Seung. 2001. Algorithms for non-negative matrix factorization. In *Advances in neural information processing systems* (pp. 556-562).
48. Sveinung Legard, George Giannoumis, Sissel Hovik, and Christina Paupini. 2019. Variation in E-Participation Schemes and Strategies: Comparative Case Study of Oslo, Madrid, and Melbourne. Proceedings of the 12[th] International Conference on Theory and Practice of Electronic Governance, 144–147.
49. Regina List, Marcel Hadeed, Rafael Schmuziger Goldzweig, and Jessica Leong Cohen. 2018. Online Participation in Culture and Politics: Towards more Democratic Societies? Council of Europe.
50. Helen Liu. 2017. Crowdsourcing Government: Lessons from Multiple Disciplines. Public Administration Review, 77(5), 656–667.
51. Ann Macintosh, and Efthimios Tambouris (Eds). 2009. *Electronic Participation: First International Conference, ePart 2009 Linz, Austria, August 31–September 4, Proceedings*. Vol. 5694. Springer.
52. Helen Margetts, Peter John, Scott Hale, and Taha Yasseri. 2015. Political turbulence: How social media shape collective action. Princeton University Press.
53. Michael Margolis, and Gerson Moreno-Riaño. 2013. The prospect of internet democracy. Ashgate Publishing.
54. Rony Medaglia, and Yang Yang. 2017. Online public deliberation in China: Evolution of interaction patterns and network homophily in the Tianya discussion forum. Information, Communication & Society, 20(5), 733–753.
55. Ines Mergel, Karl Rethemeyer, and Kimberley Isett. 2016. Big data in public affairs. *Public Administration Review* 76, no. 6: 928-937.
56. Rada Mihalcea, and Paul Tarau. 2004. Textrank: Bringing order into text. In *Proceedings of the 2004 conference on empirical methods in natural language processing* (pp. 404-411).
57. Geoff Mulgan. 2018. Big mind: How collective intelligence can change our world. Princeton University Press.
58. Cataldo Musto, Giovanni Semeraro, Pasquale Lops, and Marco de Gemmis. 2015. CrowdPulse: A framework for real-time semantic analysis of social streams. Information Systems, 54, 127–146.
59. Azadeh Nematzadeh, Giovanni Luca Ciampaglia, Yong Yeol Ahn, and Alessandro Flammini. 2019. Information overload in group communication: From conversation to cacophony in the Twitch chat. Royal Society Open Science, 6(10), 191412.
60. Jacob Nielsen. 1994. Usability engineering. Morgan Kaufmann.
61. Beth Simone Noveck. 2018. Crowdlaw: Collective intelligence and lawmaking. *Analyse & Kritik* 40, no. 2: 359-380.
62. Sylvia Octa Putri. 2020. The 2018 United Nations E-government Survey.
63. Katrin Oddsdóttir. 2014. Iceland: The Birth of the World's First Crowd-Sourced Constitution? Cambridge International Law Journal, 3(4), 1207–1220.
64. Pentti Paatero, and Unto Tapper. 1994. Positive matrix factorization: A non-negative factor model with optimal utilization of error estimates of data values. Environmetrics, 5: 111-126. doi:10.1002/env.3170050203
65. Eleni Panopoulou, Efthimios Tambouris, and Konstantinos Tarabanis. 2010. eParticipation initiatives in Europe: learning from practitioners. In *International conference on electronic participation*, pp. 54-65. Springer, Berlin, Heidelberg.
66. Mirko Pečarič. 2017. Can a group of people be smarter than experts? The Theory and Practice of Legislation, 5(1), 5–29.
67. Fabian Pedregosa, Gael Varoquaux, Alexandre Gramfort, Vincent Michel, Bertrand Thirion, Oliver Grisel, et al. 2011. Scikit-learn: Machine learning in Python. *the Journal of machine Learning research*, *12*, 2825-2830.
68. Jeffrey Pennington, Richard Socher, and Christopher Manning. 2014. Glove: Global vectors for word representation. In *Proceedings of the 2014 conference on empirical methods in natural language processing (EMNLP)* (pp. 1532-1543).
69. Oren Perez. 2008. Complexity, Information Overload, and Online Deliberation. ISJLP, 5, 43.
70. John Prpić, Araz Taeihagh, and James Melton. 2015. The fundamentals of policy crowdsourcing. *Policy & Internet* 7, no. 3: 340-361.
71. Peng Qi, Timothy Dozat, Yuhao Zhang, and Christopher D. Manning. 2019. Universal Dependency Parsing from Scratch. In Proceedings of the CoNLL 2018 Shared Task: Multilingual Parsing from Raw Text to Universal Dependencies, pp. 160-170.

72. Peter Gordon Roetzel. 2019. Information overload in the information age: a review of the literature from business administration, business psychology, and related disciplines with a bibliometric approach and framework development. *Business Research* 12, no. 2: 479-522.
73. Herbert Simon. 1957. A behavioral model of rational choice. Models of man, social and rational: Mathematical essays on rational human behavior in a social setting, 241-260.
74. Julie Simon, Theo Bass, Victoria Boelman, and Geoff Mulgan. 2017. Digital democracy: the tools transforming political engagement. *NESTA, UK, England and Wales* 1144091.
1. Yves Sintomer, Carsten Herzberg, and Anja Röcke. 2008. Participatory budgeting in Europe: Potentials and challenges. International Journal of Urban and Regional Research, 32(1), 164–178.
75. Min Song, Meen Chul Kim, and Yoo Kyung Jeong. 2014. Analyzing the political landscape of 2012 Korean presidential election in Twitter. IEEE Intelligent Systems, 29(2), 18–26.
76. Cass Sunstein. 2006. Infotopia: How many minds produce knowledge. Oxford University Press.
77. James Surowiecki. 2005. The wisdom of crowds. Anchor.
78. Silvia Suteu. 2015. Constitutional conventions in the digital era: Lessons from Iceland and Ireland. *BC Int'l & Comp. L. Rev.* 38: 251.
79. Don Tapscott, and Anthony Williams. 2006. Wikinomics: How mass collaboration changes everything. Penguin.
80. Mariona Taulé, Maria Martí and Marta Recasens. 2008. AnCora: Multilevel Annotated Corpora for Catalan and Spanish.
81. Maarja Toots. 2019. Why E-participation systems fail: The case of Estonia's Osale. ee. *Government Information Quarterly*.
82. Long Tran-Thanh, Sebastian Stein, Alex Rogers, and Nicholas Jennings. 2014. Efficient crowdsourcing of unknown experts using bounded multi-armed bandits. Artificial Intelligence, 214, 89–111.
83. Aspasia Tsaoussi. 2014. Bounded Rationality. In J. Backhaus (Ed.), Encyclopedia of Law and Economics (pp. 1–6). Springer New York.
84. Economist Intelligence Unit. 2016. Democracy index 2015: Democracy in an age of anxiety.
85. Jos Verhulst, and Arjen Nijeboer. 2007. Direct democracy: Facts and arguments about the introduction of initiative and referendum. *Democracy International, Brussels*.
86. Thomas Vollmann. 1991. Cutting the Gordian knot of misguided performance measurement. Industrial Management & Data Systems.
87. Ariadne Vromen. 2017. Digital citizenship and political engagement. In *Digital Citizenship and Political Engagement*, pp. 9-49. Palgrave Macmillan, London.
88. Bill Wanlund. 2020. Global Protest Movements. CQ Researcher, 1–58.
89. Mijail Zolotov, Tiago Oliveira, and Sven Casteleyn. 2018. E-participation adoption models research in the last 17 years: A weight and meta-analytical review. *Computers in Human Behavior* 81: 350-365.